\documentclass[10pt,twocolumn,letterpaper]{article}

\usepackage{cvpr}
\usepackage{times}
\usepackage{epsfig}
\usepackage{graphicx}
\usepackage{amsmath}
\usepackage{amssymb}

\usepackage{latexsym}
\usepackage{amsbsy}
\usepackage{epstopdf}
\usepackage{verbatim}
\usepackage{color}
\usepackage{transparent}
\usepackage{algorithm}
\usepackage{algpseudocode}
\usepackage{pifont}
\usepackage[toc,page]{appendix}

\DeclareMathOperator*{\argmax}{argmax}
\DeclareMathOperator*{\argmin}{argmin}

\renewcommand{\vec}[1]{\mathbf{#1}}

\usepackage[colorinlistoftodos,prependcaption,textsize=small]{todonotes}

\usepackage[pagebackref=true,breaklinks=true,letterpaper=true,colorlinks,bookmarks=false]{hyperref}

\cvprfinalcopy % *** Uncomment this line for the final submission

\def\cvprPaperID{1852} % *** Enter the CVPR Paper ID here

\ifcvprfinal\fi
\begin{document}

%%%%%%%%% TITLE
\title{A Hierarchical Pose-Based Approach to Complex Action Understanding Using
Dictionaries of Actionlets and Motion Poselets}

\author{Ivan Lillo\\
\normalsize P. Universidad Catolica de Chile\\
\normalsize Santiago, Chile\\
{\tt\small ialillo@uc.cl}
\and
Juan Carlos Niebles\\
\normalsize Stanford University, USA\\
\normalsize Universidad del Norte, Colombia\\
{\tt\small jniebles@cs.stanford.edu}
\and
Alvaro Soto\\
\normalsize P. Universidad Catolica de Chile\\
\normalsize Santiago, Chile\\
{\tt\small asoto@ing.uc.cl}
}

\maketitle
%\thispagestyle{empty}

%%%%%%%%% ABSTRACT
\begin{abstract}
In this paper, we introduce a new hierarchical model for human action
recognition using body joint locations.
Our model can categorize complex actions in videos,
and perform spatio-temporal annotations of the atomic
actions that compose the complex action being performed.
That is, for each atomic action, the model generates temporal action annotations by
estimating its starting and ending times, as well as,
spatial annotations by inferring the human body parts that are involved
in executing the action. Our model includes three key 
novel properties:
(i) it can be trained with no spatial
supervision, as it can automatically discover active body parts
from temporal action annotations only;
(ii) it jointly learns flexible representations for motion poselets and 
actionlets that encode the
visual variability of body parts and atomic actions;
(iii) a mechanism to discard idle or non-informative body 
parts which increases its robustness to common pose estimation errors. We 
evaluate the performance of our method using multiple
action recognition benchmarks. Our model consistently outperforms baselines
and state-of-the-art action recognition methods.

\end{abstract}

%%%%%%%%% BODY TEXT
\section{Introduction} \label{sec:introduction}  
\begin{figure}[t]
\begin{center} \label{fig:frontfigure}
\includegraphics[width=0.98\linewidth]{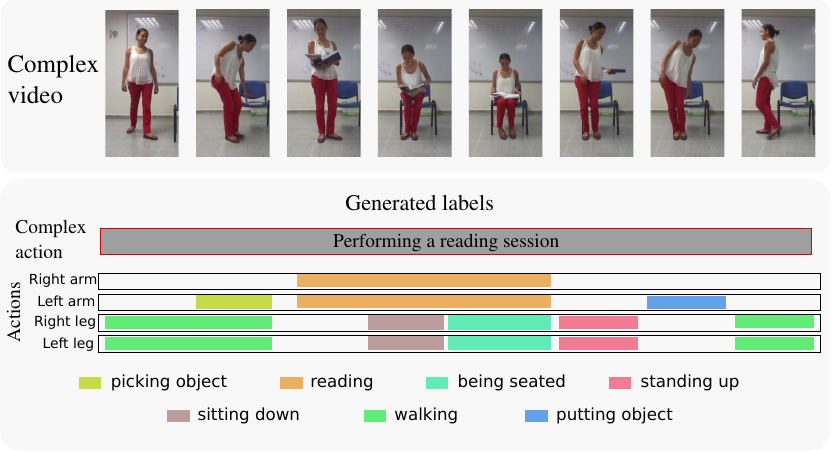}
%\includegraphics[height=0.28\linewidth]{jc/fig1_spatial.pdf}
%&
%\includegraphics[height=0.28\linewidth]{jc/fig1_temporal.pdf}\\
%(a) Spatial & (b) Temporal composition
\vspace{-0.5cm}
\end{center}
\caption{\footnotesize
Sample frames from a video sequence featuring a complex action. 
Our method is able to identify the global complex action, as well as, the temporal and 
spatial span of meaningful actions (related to \emph{actionlets}) and local body part 
configurations (related to \emph{motion poselets}).}
%\vspace{-0.5cm}
\end{figure}

Human action recognition in video is a key technology for a
wide variety of applications, such as smart surveillance, human-robot interaction,
and video search. Consequently, it has received wide attention
in the computer vision community with a strong focus on recognition of single
actions in short video sequences
\cite{Aggarwal2011,Poppe2010,vishwakarma2013survey,weinland2011survey}.
As this area
evolves, there has been an increasing interest to develop more
flexible models that can extract useful knowledge from longer video 
sequences, featuring
multiple concurrent or sequential actions, which we refer to as
\textit{complex actions}.
Furthermore, to facilitate tasks such as
video tagging or retrieval, it is important to design models
that can identify the spatial and temporal spans of each relevant
action. As an example, Figure \ref{fig:frontfigure} illustrates a potential 
usage
scenario, where an input video featuring a complex action is automatically
annotated by identifying its underlying atomic actions and corresponding 
spatio-temporal spans.

A promising research direction for reasoning about complex human actions is to
explicitly incorporate body pose representations. In effect, as noticed long 
ago, body poses are highly informative to discriminate 
among human actions \cite{Johansson:1973}. Similarly, recent works have also 
demonstrated the relevance of explicitly 
incorporating body pose information in action recognition models 
\cite{Jhuang2013,Wang2013}. While human body pose estimation from color images 
remains 
elusive, the emergence of accurate and cost-effective RGBD 
cameras has enabled the development of robust 
techniques to identify body joint locations and to infer body poses 
\cite{Shotton:EtAl:11}. 

In this work, we present a new 
pose-based approach to recognizing
and provide detailed information about complex human actions in RGBD 
videos. Specifically, given a video featuring a complex action, our
model can identify the complex action occurring in the video, as well 
as, the set of atomic actions that compose this complex action. 
Furthermore, for each atomic action, the model is also able to generate 
temporal 
annotations by estimating its starting and 
ending times, and spatial annotations by inferring the body
parts that are involved in the action execution.

To achieve this, we propose a hierarchical compositional model that
operates at three levels of abstraction: body poses, atomic actions, and 
complex actions. At the level of body poses, our model learns a dictionary 
that captures relevant 
spatio-temporal configurations of body parts. We refer to the components 
of this dictionary as \textit{motion 
poselets} \cite{Bourdev:EtAl:2010, Tao2015}. At the level of atomic actions, 
our model learns a dictionary that captures the main 
modes of variation in the execution of each action. We refer to the components 
of this dictionary as \textit{actionlets} \cite{Wang2012}. Atoms in both 
dictionaries are given by linear classifiers that are jointly learned by 
minimizing an energy 
function that constraints compositions among \textit{motion poselets} and 
\textit{actionlets}, as 
well as, their spatial and temporal relations. While 
our 
approach can be extended to more general cases, here we focus on
modeling atomic actions that can be characterized by the body motions of a 
single actor, such as running, drinking, or eating.  

Our model introduces several contributions with respect to
prior work \cite{Lillo2014,Tao2015,Wang2012,Wang2008}.
First, it presents a novel formulation based 
on a structural latent SVM model \cite{Yu:Joachims:2010} and an initialization 
scheme based on self-pace learning \cite{Kumar:EtAl:2010}. These provide an 
efficient and robust mechanism to infer, at test and training time, action 
labels for each detected motion poselet, as well as, their temporal and spatial 
span. Second, it presents a multi-modal approach that
trains a group of actionlets for each atomic action.
This provides a robust method to capture relevant 
intra-class variations in action execution. Third, it incorporates a \textit{garbage 
collector} mechanism that identifies and discards idle or non-informative 
spatial areas of the input videos. This provides an effective method to process 
long video sequences. Finally, we provide empirical evidence indicating that the 
integration of the previous contributions in a single hierarchical model, 
generates a highly informative and accurate solution that outperforms 
state-of-the-art approaches.

%In the rest of the paper we review related previous work
%(Sec.~\ref{sec:related_work}),
%describe the proposed method
%(Sec.~\ref{sec:model}),
%present qualitative and quantitative experiments
%on benchmark datasets
%(Sec.~\ref{sec:experiments});
%and close with conclusions (Sec.~\ref{sec:conclusions}).

%\input{introduction_IL}+++
%
%----------------------------------------------------------------
%
\section{Related Work} \label{sec:related_work}  There is a large body of work on human activity recognition in the computer
vision literature
\cite{Aggarwal2011,Poppe2010,vishwakarma2013survey,weinland2011survey}.
We focus on recognizing human actions and
activities from videos using pose-based representations and review in the
following some of the most relevant previous work.

The idea of using human body poses and configurations as an important
cue for recognizing human actions has been explored recurrently,
as poses provide strong cues on the actions
being performed.
Initially, most research focused on pose-based action recognition in color
videos \cite{Feng2002, Thurau2008}.
But due to the development of pose estimation
methods on depth images\cite{Shotton:EtAl:11}, there has been recent interest in
pose-based action recognition from RGBD videos
\cite{Escorcia2012, Hu2015, Vemulapalli2014}.
Some methods have tackled the problem of jointly recognizing
actions and poses in videos \cite{Nie2015} and still images \cite{Yao2010},
with the hope to create positive feedback by solving both tasks simultaneously.

One of the most influential pose-based representations in the literature
is Poselets, introduced by Bourdev and Malik \cite{Bourdev2009}.
Their representation relies on the construction of a large set of
frequently occurring poses, which is used to represent the pose space in a
quantized, compact and discriminative manner.
Their approach has been applied to action recognition in still
images \cite{maji2011action},
as well as in videos \cite{Tao2015, Wang2014,Zanfir2013}.

%\paragraph{Representation} Poselets.
%Moving Poselets \cite{Tao2015}.
%Dynamic Poselets \cite{Wang2014}.
%Moving Pose, a descriptor for action recognition \cite{Zanfir2013}

Researchers have also explored the idea of fusing pose-based cues with
other types of visual descriptors. For example, Cheron \etal \cite{Cheron2015}
introduce P-CNN as a framework for incorporating pose-centered
CNN features extracted from optical flow and color.
In the case of RGBD videos, researchers have proposed the fusion
of depth and color features \cite{Hu2015, Kong2015}.
In general, the use of multiple types of features helps to disambiguate some
of the most similar actions.

Also relevant to our framework are hierarchical models for action
recognition. In particular, the use of latent variables as an intermediary
representation in the internal layers of the model can be a powerful
tool to build discriminative models and meaningful representations
\cite{Hu2014, Wang2008}. An alternative is to learn hierarchical models
based on recurrent neural networks \cite{YongDu2015}, but they tend to lack
interpretability in their internal layers and require very large amounts
of training data to achieve good generalization.

%\paragraph{Models} Latent variable models for action recognition.
%Latent models for activity recognition \cite{Hu2014}.
%\paragraph{Models} Neural Networks. RNN \cite{YongDu2015}

While most of the previous work have focused on recognizing single and isolated
simple actions, in this paper we are interested in the recognition
of complex, composable and concurrent actions
and activities. In this setting, a person may be executing multiple actions
simultaneously, or in sequence, instead of performing each action in isolation.
An example of these is the earlier work of Ramanan and Forsyth \cite{Ramanan2003},
with more recent approaches by Yeung \etal \cite{Yeung2015} and
Wei \etal \cite{Wei2013}.
Another recent trend aims at fine-grained detection of actions performed in sequence
such as those in a cooking scenario \cite{Rohrbach2012, Lan2015}.

We build our model upon several of these ideas in the literature. Our method
extends the state-of-the-art by introducing a model that can perform
detailed annotation of videos during testing time but only requires weak
supervision at training time. While learning can be done with
reduced labels, the hierarchical structure of poselets and actionlets combined
with other key mechanisms enable our model to achieve improved
performance over competing methods in several evaluation benchmarks.

%
%-------------------------------------------------------------------------
%
\section{Model Description} \label{sec:model} \begin{figure}[tb]
\begin{center}
\includegraphics[width=0.95\linewidth]{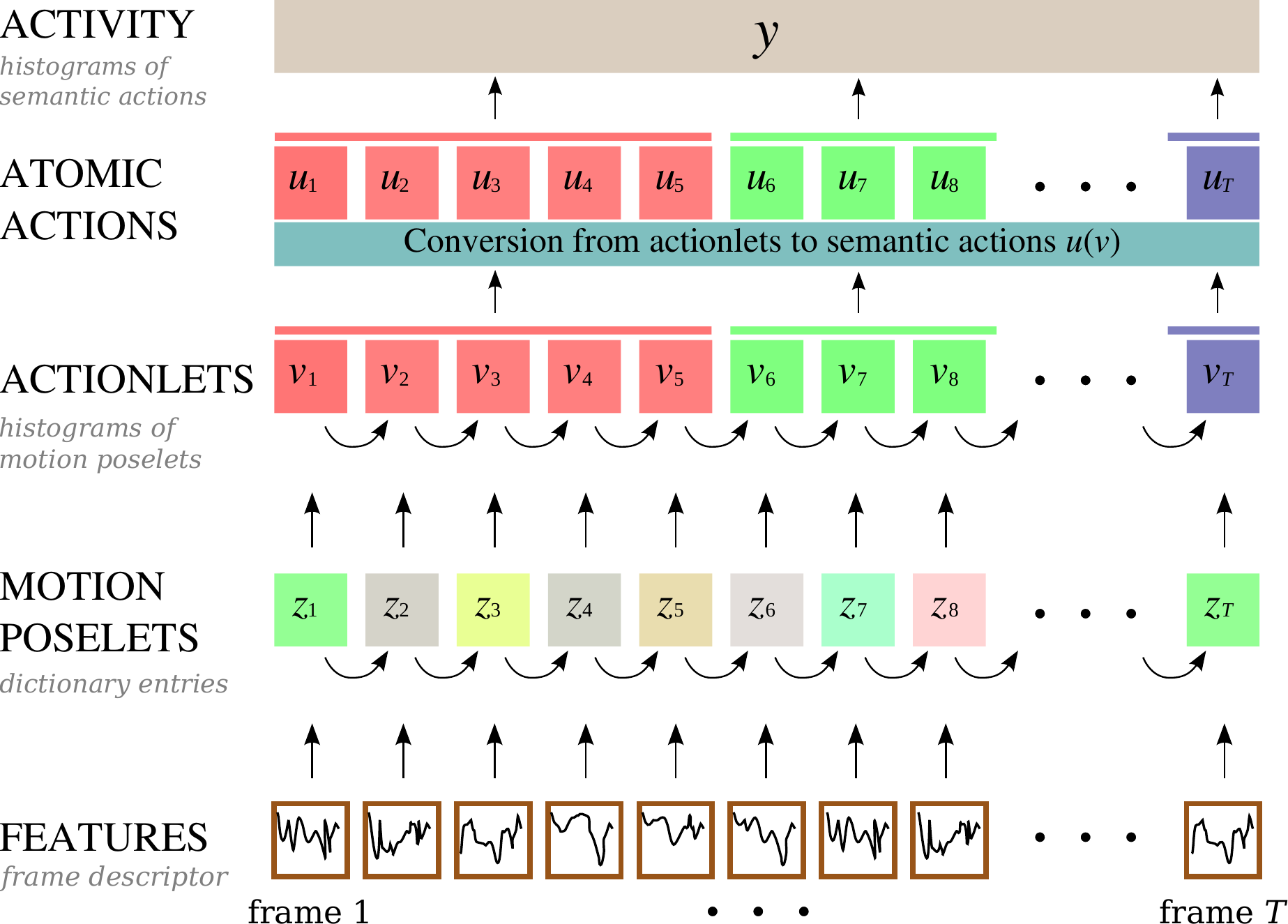}
\vspace{-2mm}
\end{center}
\caption{\footnotesize
Graphical representation of our discriminative hierarchical model for
recognition of complex human actions.
At the top level, activities are represented as compositions of atomic actions that are inferred at
the intermediate level. These actions are, in turn, compositions of poses at the
lower level, where pose dictionaries are learned from data. Our model also learns
temporal transitions between consecutive poses and actions.}
\label{fig:overview}
%\vspace{-4mm}
\end{figure}

In this section, we introduce our model for pose-based recognition of complex 
human actions. Our goal is to build a model with the capability of 
annotating input videos with the actions being performed, automatically identifying the parts of the body 
that are involved in each action (spatial localization) along with the temporal 
span of each action (temporal localization).
As our focus is on
concurrent and composable activities, we would also like to encode multiple
levels of abstraction, such that we can reason about poses, actions, and their
compositions. Therefore, we develop a hierarchical compositional framework for 
modeling and recognizing complex human actions.

One of the key contributions of our model is its capability to spatially 
localize the body regions that are involved in the execution of each action, 
\emph{both at training and testing time}. Our training process does not 
require careful spatial annotation and localization of actions in the training 
set; instead, it uses temporal annotations of actions only.
At test time, it can discover the spatial and temporal span, as well as, 
the specific configuration of the main body regions executing each action.
We now introduce the components of our model and the training
process that achieves this goal.

\begin{figure}[tb]
\begin{center}
\includegraphics[width=0.8\linewidth]{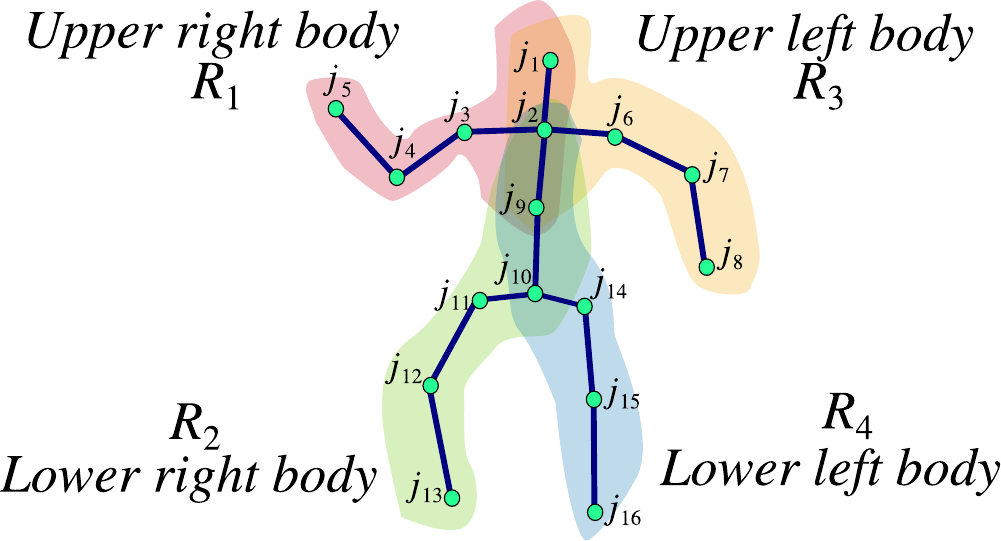}
\vspace{-4mm}
\end{center}
\caption{
\footnotesize
Skeleton representation used for splitting the human body into a set of 
spatial regions.}
\label{fig:skeleton_limbs_regions}
%\vspace{-4mm}
\end{figure}

\subsection{Body regions}
We divide the body pose into $R$ fixed spatial regions and independently compute 
a pose feature vector for each region. Figure \ref{fig:skeleton_limbs_regions} 
illustrates the case when $R = 4$ that we use in all our experiments. Our body 
pose feature vector consists of the concatenation of two descriptors. At frame 
$t$ and region $r$, a descriptor $x^{g}_{t,r}$ encodes geometric information 
about the spatial configuration of body joints, and a descriptor $x^{m}_{t,r}$ 
encodes local motion information around each body joint position.
%Following 
%\cite{Lillo2014},
We use the geometric descriptor from \cite{Lillo2014}:
we construct six segments that connect pairs of joints at each
region\footnote{Arm segments: wrist-elbow, elbow-shoulder, shoulder-neck, wrist-shoulder, wrist-head, and neck-torso;  Leg segments: ankle-knee, knee-hip, hip-hip center, ankle-hip, ankle-torso and hip center-torso}
and compute 15 angles between those segments.
Also, three angles are calculated between a plane formed by three
segments\footnote{Arm plane: shoulder-elbow-wrist;  Leg plane: hip-knee-ankle} and 
the remaining three non-coplanar segments, totalizing an 18-D geometric descriptor (GEO) for every region.
% Following \cite{WangCVPR2011},
Our motion descriptor is based on tracking motion trajectories of key points
\cite{WangCVPR2011}, which in our case coincide with body joint positions.
We extract a HOF descriptor
using 32x32 RGB patches centered at the joint location for a temporal window of 15 
frames. At each joint location, this produces a 108-D descriptor,  
which we concatenate across all joints in each a region to obtain our motion descriptor. Finally, 
we apply PCA to reduce the dimensionality of our concatenated motion descriptor
to 20. The final descriptor is the concatenation of the geometric and 
motion descriptors, $x_{t,r} = [x_{t,r}^g ; x_{t,r}^m]$.
%into a 20-dimensional vector.
%, keeping the dimensionality of our final descriptor
%relatively low.

\subsection{Hierarchical compositional model}

We propose a hierarchical compositional model that spans three semantic 
levels. Figure \ref{fig:overview} shows a schematic of our model. At the 
top level, our model assumes that each input video has a single complex action 
label $y$. Each complex action is composed of a 
temporal and spatial arrangement of atomic actions with labels $\vec{u}=[u_1,\dots,u_T]$, $u_i \in \{1,\dots,S\}$.
In turn, each atomic action consists of several non-shared \emph{actionlets}, which correspond to representative sets of pose configurations for action identification, modeling the multimodality of each atomic action.
We capture actionlet assignments in $\vec{v}=[v_1,\dots,v_T]$, $v_i \in \{1,\dots,A\}$.
Each actionlet index $v_i$ corresponds to a unique and known actomic action label $u_i$, so they are related by a mapping $\vec{u} = \vec{u}(\vec{v})$. At the 
intermediate level, our model assumes that each actionlet is composed of a 
temporal arrangement of a subset from $K$ body poses, encoded in $\vec{z} = [z_1,\dots,z_T]$, $z_i \in \{1,\dots,K\}$,
where $K$ is a hyperparameter of the model.
These subsets capture pose geometry and local motion, so we call them \emph{motion poselets}.
Finally, at the bottom level, our model identifies motion poselets 
using a bank of linear classifiers that are applied to the incoming frame 
descriptors.

%Similarly to previous works \cite{Lillo2014, Taralova:EtAl:2014}
We build each layer of our hierarchical model on top of BoW 
representations of labels. To this end, at the bottom level of our hierarchy, and for 
each body region, we learn a dictionary of motion poselets. Similarly, at the mid-level of our hierarchy, we learn a dictionary of actionlets, using the BoW representation of motion poselets as inputs. At each of these levels, 
spatio-temporal activations of the respective dictionary words are used 
to obtain the corresponding histogram encoding the BoW representation. 
The next two sections provide
details on the process to represent and learn the dictionaries of motion 
poselets and actionlets. Here we discuss our
integrated hierarchical model.

We formulate our hierarchical model using an energy function.
Given a video of $T$ frames corresponding to complex action $y$ encoded by descriptors $\vec{x}$, with the label vectors $\vec{z}$ for motion poselets,
$\vec{v}$ for actionlets and $\vec{u}$ for atomic actions, we
define an energy function for a video as:
\small
\begin{align}\label{Eq_energy}
%\begin{split}
E(\vec{x},&\vec{v},\vec{z},y) =  E_{\text{motion poselets}}(\vec{z},\vec{x}) \nonumber \\&+ E_{\text{motion poselets BoW}}(\vec{v},\vec{z}) + 
E_{\text{atomic actions BoW}}(\vec{u}(\vec{v}),y) \nonumber \\ 
& + E_{\text{motion poselets transition}}(\vec{z}) + E_{\text{actionlets 
transition}}(\vec{v}).
%\end{split}
\end{align}
\normalsize
Besides the BoW representations and motion poselet classifiers
described above, Equation (\ref{Eq_energy}) includes
two energy potentials that encode information related to
temporal
transitions between pairs of motion poselets ($E_{\text{motion poselets 
transition}}$) and 
actionlets ($E_{\text{actionlets transition}}$). 
%
%Considering the BoW representations and linear classifiers to identify motion 
%poselets, 
The energy potentials are given by:
{\small
\begin{align}
%\begin{equation}
\label{eq:motionposelets}
&E_{\text{mot. poselet}}(\vec{z},\vec{x})  = \sum_{r,t}  \left[ \sum_{k} {w^r_k}^\top 
x_{t,r}\delta_{z_{(t,r)}}^{k} + \theta^r \delta_{z_{(t,r)}}^{K+1}\right] \\
%\end{equation}
%\begin{equation}
&E_{\text{mot. poselet BoW}}(\vec{v},\vec{z})  = \sum_{r,a,k} {\beta^r_{a,k}}\delta_{v_{(t,r)}}^{a}\delta_{z_{(t,r)}}^{k}\\
%\end{equation}
%\begin{equation}
\label{eq:actionlets_BoW} 
&E_{\text{atomic act. BoW}}(\vec{u}(\vec{v}),y) =\sum_{r,s} {\alpha^r_{y,s}}\delta_{u(v_{(t,r)})}^{s} \\
%\end{equation}
%\begin{equation}
&E_{\text{mot. pos. trans.}}(\vec{z})  = 
\sum_{r,k_{+1},k'_{+1}} \eta^r_{k,k'} 
\sum_{t} \delta_{z_{(t-1,r)}}^{k}\delta_{z_{(t,r)}}^{k'} \\
%\end{equation}
%\begin{equation} 
\label{eq:actionletstransition}
&E_{\text{acttionlet trans.}}(\vec{v})  =\sum_{r,a,a'} \gamma^r_{a,a'} 
\sum_{t} 
\delta_{v_{(t-1,r)}}^{a}\delta_{v_{(t,r)}}^{a'} 
%\end{equation}
\end{align}
}

Our goal is to 
maximize $E(\vec{x},\vec{v},\vec{z},y)$, and obtain the 
spatial and temporal arrangement 
of motion poselets $\vec{z}$ and actionlets $\vec{v}$, as well as, the underlying 
complex action $y$. %\JC{Revisar}

In the previous equations, we use $\delta_a^b$ to indicate the Kronecker delta function $\delta(a = b)$, and use indexes $k \in \{1,\dots,K\}$ for motion poselets, $a \in \{1,\dots,A\}$ for actionlets, and $s \in \{1,\dots,S\}$ for atomic actions.
In the energy term for motion poselets,
$w^r_k$ are a set of $K$ linear pose classifiers applied to frame 
descriptors $x_{t,r}$, according to the label of the latent variable $z_{t,r}$. 
Note that there is a special label $K+1$; the role of this label will be 
explained in Section \ref{subsec:garbage_collector}.
In the energy potential associated to 
the BoW representation for motion poselets, $\vec{\beta}^r$ denotes a set of $A$ 
mid-level classifiers, whose inputs are histograms of motion 
poselet labels at those frame annotated as actionlet $a$. At the highest level, 
$\alpha^r_{y}$ is a linear classifier associated with complex action $y$, whose 
input is the histogram of atomic action labels,
which are related to actionlet assignments by the mapping function $\vec{u}(\vec{v})$. Note that all classifiers 
and labels here correspond to a single region $r$. We add the contributions of all 
regions to compute the global energy of the video. The transition terms act as
linear classifiers $\eta^r$ and $\gamma^r$ over histograms of temporal transitions of motion poselets 
and temporal transitions of actionlets respectively. As we have a special label $K+1$ for motion poselets, the summation index
%are denoted with a ``+1'' subscript to indicate
$k_{+1}$ indicates the interval $\lbrack 1,\dots,K+1 \rbrack$.

\subsection{Learning motion poselets}
In our model, motion poselets are learned by treating them as latent variables  
during training. Before training, we fix the number of motion poselets per region to $K$.
In every region $r$, we learn an independent
set of pose classifiers $\{w^r_k\}_{k=1}^K$, initializing the motion poselet 
labels using the $k$-means algorithm. We learn pose classifiers, 
actionlets and complex actions classifiers jointly, allowing the model to discover 
discriminative motion poselets useful to detect and recognize complex actions. 
As shown in previous work, jointly learning linear
classifiers to identify body parts and atomic actions improves recognition 
rates \cite{Lillo2014,Wang2008}, so here we follow a similar hierarchical 
approach, and integrate learning
of motion poselets with the learning of actionlets.

\subsection{Learning actionlets}
\label{sec:learningactionlets}
A single linear classifier does not offer enough flexibility to identify atomic 
actions that exhibit high visual variability. As an example, the atomic action 
``open'' can be associated with ``opening a can'' or ``opening a 
book'', displaying high variability in action execution. Consequently, we 
augment our hierarchical model including multiple classifiers to 
identify different modes of action execution. 

Inspired by \cite{Raptis2012}, we use the \emph{Cattell's Scree test} to
find a suitable number of actionlets to model each atomic 
action. Specifically, using the atomic action labels, we compute a descriptor 
for every video interval using 
normalized histograms of initial pose labels obtained with $k$-means. Then, for a particular atomic action 
$s$, we compute the eigenvalues $\lambda(s)$ of the affinity matrix of the 
atomic action descriptors, which is build using $\chi^2$ distance. For each 
atomic action 
$s \in \{1,\dots,S\}$, we find the number of actionlets $G_s$ as $G_s = 
\argmin_i {\lambda(s)}_{i+1}^2 / (\sum_{j=1}^i {\lambda(s)}_j) + c\cdot i$, with $c=2\cdot 
10^{-3}$. Finally, we cluster the descriptors from each atomic 
action $s$ running $k$-means with $k = G_s$. This scheme generates 
a set of non-overlapping actionlets to model each single atomic 
action. In our experiments, we notice that the number of actionlets used to 
model each atomic action varies typically from 1 to 8.

To transfer the new labels to the model, we define $u(v)$ as a function that
maps from actionlet label $v$ to the corresponding atomic action label 
$u$. A dictionary of actionlets provides a richer representation for actions, 
where several actionlets will map to a single atomic action. This behavior 
resembles a max-pooling operation, where at inference time we will choose the 
set of actionlets that best describe the performed actions in the video, keeping 
the semantics of the original atomic action labels.

\subsection{A garbage collector for motion poselets}
\label{subsec:garbage_collector}
While poses are highly informative for action recognition, an input video 
might contain irrelevant or idle zones, where the underlying poses are noisy 
or non-discriminative to identify the actions being performed in the video. As 
a result, low-scoring motion poselets could degrade the pose classifiers during 
training, decreasing their performance. To deal with this problem, we include in 
our model a \emph{garbage collector} mechanism for motion poselets. This 
mechanism operates by assigning all low-scoring motion poselets to
the $(K+1)$-th pose dictionary entry. These collected poses are 
associated with a learned score lower than $\theta^r$, as in Equation 
(\ref{eq:motionposelets}). Our experiments show that this mechanism leads 
to learning more discriminative motion poselet classifiers.

\subsection{Learning} \label{subsec:learning}
\textbf{Initial actionlet labels.} An important step in the training process is
the initialization of latent variables. This is a challenging due to the lack
of spatial supervision: at each time instance, the available atomic actions can be associated with 
any of the $R$ body regions.
We adopt the machinery of 
self-paced 
learning \cite{Kumar:EtAl:2010} to provide a suitable solution and 
formulate the association between actions and body regions as an 
optimization problem. We constrain this optimization using two structural 
restrictions:
i) atomic actions intervals must not overlap in the same region, and 
ii) a labeled atomic action must be present at least in one region. We 
formulate the labeling 
process as a binary Integer Linear Programming (ILP) problem, where we define 
$b_{r,q}^m=1$ when action interval $q \in \{1,\dots,Q_m\}$ is active in region 
$r$ of video $m$; and $b_{r,q}^m=0$ otherwise. Each action interval $q$ is 
associated with a single atomic action. We assume that we have initial 
motion poselet labels
$z_{t,r}$ in each frame and region.
We describe the action interval $q$ and region $r$ using 
the histogram $h_{r,q}^m$ of motion poselet labels. We can find 
the correspondence between action intervals and regions using a formulation 
that resembles the operation of$k$-means, but using the
structure of the problem to constraint the labels:
\small
\begin{equation}
\begin{split}
\text{P1}) \quad \min_{b,\mu} &\sum_{m=1}^M  \sum_{r=1}^R \sum_{q=1}^{Q_m}  b_{r,q}^m 
d( h_{r,q}^m - \mu_{a_q}^r) -\frac{1}{\lambda} b_{r,q}^m\\ 
 \text{s.t.} 
\quad 
& \sum_{r=1}^R b_{r,q}^m \ge 1\text{, }\forall q\text{, }\forall m \\ 
%& \sum_{q=1}^{Q_m} v_{r,q}^m \le t_m \\ 
& b_{r,q_1}^m + b_{r,q_2}^m \le 1 \text{ if } q_1\cap q_2 \neq \emptyset 
\text{, 
}\forall r\text{, }\forall m\\  
& b_{r,q}^m \in \{0,1\}\text{, }\forall q\text{, }\forall{r}\text{, }\forall m
\end{split}
\end{equation}
with
\begin{equation}
d( h_{r,q}^m - \mu_{a_q}^r) = \sum_{k=1}^K (h_{r,q}^m[k] - 
\mu_{a_q}^r[k])^2/(h_{r,q}^m[k] +\mu_{a_q}^r[k]).
\end{equation}
\normalsize
Here, $\mu_{a_q}^r$ are the means of the descriptors with action 
label $a_q$ within region $r$. We solve $\text{P1}$ iteratively using a block coordinate 
descending scheme, alternating between solving $b_{r,q}^m$ with $\mu_{a}^r$ 
fixed, which has a trivial solution; and then fixing $\mu_{a}^r$ to solve 
$b_{r,q}^m$, relaxing $\text{P1}$ to solve a linear program. Note that the second term 
of the objective function in $\text{P1}$ resembles the objective function of 
\emph{self-paced} learning \cite{Kumar:EtAl:2010}, managing the balance between 
assigning a single region to every action or assigning all possible regions to 
the respective action interval.  

\textbf{Learning model parameters.}
We formulate learning the model parameters as a Latent Structural SVM
problem \cite{Yu:Joachims:2010}, with latent variables for motion
poselets $\vec{z}$ and actionlets $\vec{v}$. We find values for parameters in 
equations 
(\ref{eq:motionposelets}-\ref{eq:actionletstransition}),
slack variables $\xi_i$, motion poselet labels $\vec{z}_i$, and actionlet labels $\vec{v}_i$, 
by solving: % the following learning problem:
{\small
\begin{equation}
\label{eq:big_problem}
\min_{W,\xi_i,~i=\{1,\dots,M\}}    \frac{1}{2}||W||_2^2 + \frac{C}{M} \sum_{i=1}^M\xi_i ,
\end{equation}}
where
{\small \begin{equation}
W^\top=[\alpha^\top, \beta^\top, w^\top, \gamma^\top, \eta^\top, \theta^\top],
\end{equation}}
and
{\small
\begin{equation} \label{eq:slags}
\begin{split}
\xi_i = \max_{\vec{z},\vec{v},y}  \{  & E(\vec{x}_i, \vec{z}, \vec{v}, y) + \Delta( (y_i,\vec{v}_i), (y, \vec{v})) \\
 & - \max_{\vec{z}_i}{ E(\vec{x}_i, \vec{z}_i, \vec{v}_i, y_i)} \}, \; \;\; i\in[1,...M].	
\end{split}
\end{equation}}
In Equation (\ref{eq:slags}), each slack variable
$\xi_i$ quantifies the error of the inferred labeling for
video $i$. We solve Equation (\ref{eq:big_problem}) iteratively using the CCCP
algorithm \cite{Yuille:Rangarajan:03}, by solving for 
latent labels $\vec{z}_i$ and $\vec{v}_i$ given model parameters $W$, 
temporal atomic action annotations (when available), and labels of complex actions occurring in 
training videos (see Section \ref{subsec:inference}). Then, we solve for 
$W$ via 1-slack formulation using Cutting Plane algorithm 
\cite{Joachims2009}. 

The role of the loss function $\Delta((y_i,\vec{v}_i),(y,\vec{v}))$ is to penalize inference errors during 
training. If the true actionlet labels are known in advance, the loss function is the same as in \cite{Lillo2014} using the actionlets instead of atomic actions:
\small \begin{equation}
\Delta((y_i,\vec{v}_i),(y,\vec{v})) = \lambda_y(y_i \ne y) + \lambda_v\frac{1}{T}\sum_{t=1}^T 
\delta({v_t}_{i} \neq v_t),
\end{equation}
\normalsize
\noindent where ${v_t}_{i}$ is the true actionlet label. If the spatial ordering of actionlets is unknown (hence the latent 
actionlet formulation), but the temporal composition is known, we can compute a 
list $A_t$ of possible actionlets for frame $t$, and include that information
on the loss function as
\small \begin{equation}
\Delta((y_i,\vec{v}_i),(y,\vec{v})) = \lambda_y(y_i \ne y) + \lambda_v\frac{1}{T}\sum_{t=1}^T 
\delta(v_t \notin A_t)
\end{equation}
\normalsize

\subsection{Inference}
\label{subsec:inference}
The input to the inference algorithm is a new video sequence with features
$\vec{x}$. The task is to infer the best complex action label $\hat y$, and to 
produce the best labeling of actionlets $\hat{\vec{v}}$ and motion poselets $\hat{\vec{z}}$.
{\small
\begin{equation}
  \hat y, \hat{\vec{v}}, \hat{\vec{z}} = \argmax_{y, \vec{v},\vec{z}} E(\vec{x}, \vec{v}, \vec{z}, y)
\end{equation}}
We can solve this by exhaustively enumerating all values of complex actions $y$, and solving for $\hat{\vec{v}}$ and $\hat{\vec{z}}$ using:
\small
\begin{equation}
\begin{split}
 \hat{\vec{v}}, \hat{\vec{z}} | y ~ =~ &   \argmax_{\vec{v},\vec{z}} ~   \sum_{r=1}^R \sum_{t=1}^T \left( \alpha^r_{y,u(v{(t,r)})} 
                  + \beta^r_{v_{(t,r)},z_{(t,r)}}\right. \\
				&\quad\quad \left.+ {w^r_{z_{(t,r)}}}^\top x_{t,r} \delta(z_{(t,r)} \le K)  + \theta^r \delta_{z_{(t,r)}}^{K+1} \right. \\ 
				& \quad\quad \left.+ \gamma^r_{v_{({t-1},r)},v_{(t,r)}} + \eta^r_{z_{({t-1},r)},z_{(t,r)}}  \vphantom{{w^r_{z_{(t,r)}}}^\top x_{t,r}} \right). \\
\end{split}
\label{eq:classify_inference}
\end{equation}
\normalsize

%
%-------------------------------------------------------------------------
%
\section{Experiments} \label{sec:experiments} Our experimental
validation focuses on evaluating two properties of our model.
First, we measure action classification accuracy on
several action recognition benchmarks.
Second, we measure the performance of our model to provide detailed 
information about atomic actions and body regions associated to the execution 
of a complex action.

We evaluate our method on four action recognition benchmarks: the MSR-Action3D 
dataset \cite{WanLi2010}, Concurrent Actions dataset \cite{Wei2013}, Composable 
Activities Dataset \cite{Lillo2014}, and sub-JHMDB \cite{Jhuang2013}. Using 
cross-validation, we set $K=100$ in Composable Activities and Concurrent Actions 
datasets, $K=150$ in sub-JHMDB, and $K=200$ in MSR-Action3D. In all datasets, we 
fix $\lambda_y = 100$ and $\lambda_u = 25$. The number of \emph{actionlets} 
to model each atomic action is estimated using the method described in Section 
\ref{sec:learningactionlets}. 
The garbage collector (GC) label $(K+1)$ is 
automatically assigned
during inference according to the learned model parameters $\theta^r$. We initialize the $20\%$ most dissimilar frames to the $K+1$ label.
In practice, at test time, the number of frames
labeled as $(K+1)$ ranges from 14\% in MSR-Action3D to 29\% in sub-JHMDB.
%all initialized with 20\% of most dissimilar frames.

Computation is fast during testing. In the Composable Activities dataset, our 
CPU implementation runs at 300 fps on a 32-core computer, while training time is 
3 days, mostly due to the massive execution of the cutting plane algorithm. 
Using Dynamic Programming, complexity to estimate labels is linear with the 
number of frames $T$ and quadratic with the number of actionlets $A$ and motion 
poselets $K$. In practice, we filter out the majority of combinations of motion 
poses and actionlets in each frame, using the 400 best combinations of 
$(k,a)$ according to the value of non-sequential terms in the 
dynamic program. Details are provided in the supplementary material.

\subsection{Classification of Simple and Isolated Actions}

As a first experiment,
we evaluate the performance of our model on the task of simple and
isolated human action recognition in the  MSR-Action3D dataset
\cite{WanLi2010}.
Although our model is tailored at recognizing complex %human
actions, this experiment verifies the performance of our model in the
simpler scenario of isolated atomic action classification.

The MSR-Action3D dataset provides pre-trimmed depth videos and estimated body poses
for isolated actors performing actions from 20
categories. We use 557 videos %in the dataset
in a similar setup to
\cite{Wang2012}, where videos from subjects 1, 3, 5, 7, 9 are used for
training and the rest for testing. Table \ref{tab:msr3d} shows that in this 
dataset our model achieves classification accuracies comparable to 
state-of-the-art methods.

\begin{table}[t]
\footnotesize
\centering
\begin{tabular}{|l|c|}
\hline
\textbf{Algorithm} & \textbf{Accuracy}\\
\hline
Our model &  93.0\% \\
%Ours, GEO+VEL, NI &  93.0\% \\
%Ours, GEO+VEL  & 91.2\% \\
\hline
L. Tao \etal \cite{Tao2015} & 93.6\% \\
C. Wang \etal \cite{Wang2013} &    90.2\% \\
Vemulapalli \etal \cite{Vemulapalli2014} & 89.5\% \\
%Lillo et al. \cite{Lillo2014} & 89.5\%\\
\hline
\end{tabular}
\caption{\footnotesize
Recognition accuracy in the MSR-Action3D 
dataset.}
%\vspace{-1mm}
\label{tab:msr3d}
\end{table}

%\paragraph{MSR-Action3D} Setup: subjects 1,3,5,7,9 for training, rest for
%testing, using all 20 action categories. This dataset is more as a proof of
%concept, that our model achieves near state-of-the-art accuracy in a standard
%dataset.
%In fact, omitting Tao el at. ICCV 2015 paper, we would achieve the
%best accuracy. BUT, they do not provide the rich annotations for testing data
%as our model. Also, we use the same initialization method to automatically
%annotate the actions in the dataset: the initialization method is integrated
%with the model and is independent of the dataset.

%\paragraph{MSR-Action3D} A very popular skeleton + Depth single action dataset.
%We use the common setup of using skeleton data from
%, using all 20 action categories. We
%use 557 videos from the dataset as proposed by \cite{Wang2012}. We use this
%dataset to show how our model performs in a standard database of single
%actions.

\subsection{Detection of Concurrent Actions}
Our second experiment evaluates the performance of our model in a concurrent
action recognition setting. In this scenario, the goal is to predict
the temporal localization of actions that may occur concurrently in a long
video. We evaluate this task on the Concurrent Actions dataset \cite{Wei2013},
which
provides 61 RGBD videos and pose estimation data annotated with 12
action categories.
We use a similar evaluation setup as proposed by the authors.
We split the dataset into training and testing sets with a 50\%-50\% ratio.
We evaluate performance by measuring precision-recall: a detected action
is declared as a true positive if its temporal overlap with the ground
truth action interval is larger than 60\% of their union, or if
the detected interval is completely covered by the ground truth annotation.

Our model is tailored at recognizing complex actions that are composed
of atomic components. However, in this scenario, only atomic actions are
provided and no compositions are explicitly defined. Therefore, we apply
a simple preprocessing step: we cluster training videos into groups
by comparing the occurrence of atomic actions within each video.
The resulting groups are used as complex actions labels in the training
videos of this dataset.
At inference time, our model outputs a single labeling per video,
which corresponds to the atomic action labeling that maximizes the energy of
our model.
Since there are no thresholds to adjust, our model produces the single
precision-recall measurement reported in Table \ref{tab:concurrent}.
Our model outperforms the state-of-the-art method in this
dataset at that recall level.

%\paragraph{Concurrent Action dataset} This dataset has 61 videos of variable
%time, some of them are very long comoared to other action datasets. The videos
%has a variable number of actions. The skeleton data and action annotations are
%provided in the dataset.
%We select randomly 50\% of videos for training and the
%rest for testing.

%We want to show using this dataset that the latent formulation achieve good
%recognition performance with respect to the model that uses this dataset. We
%can show also new annotations in this dataset, corresponding to the regions
%that are annotated with the actions.

\begin{table}[tb]
\footnotesize
\centering
\begin{tabular}{|l|c|c|}
\hline
\textbf{Algorithm} & \textbf{Precision} & \textbf{Recall}\\
\hline
%Ours, GEO+VEL, NI, VL-KM-ST &  92.3\% & 0.81\% \\
Our full model &  0.92 & 0.81 \\
\hline
Wei et al. \cite{Wei2013} & 0.85 & 0.81 \\
\hline
\end{tabular}
\caption{
\footnotesize
Recognition accuracy in the Concurrent Actions dataset. }
\label{tab:concurrent}
%\vspace{-1mm}
\end{table}
 
\subsection{Recognition of Composable Activities}
In this experiment, we evaluate the performance of our model to recognize complex 
and composable human actions. In the evaluation, we use the Composable 
Activities dataset \cite{Lillo2014},
which provides 693 videos of 14 subjects performing 16 activities.
Each activity is a spatio-temporal composition of atomic actions.
The dataset provides a total of 26 atomic actions that are shared across
activities. We train our model using two levels of supervision during training:
i) spatial annotations that map body regions to the execution of each action are made available
ii) spatial supervision is not available, and therefore the labels $\vec{v}$ to assign spatial regions to actionlets 
are treated as latent variables.

Table \ref{tab:composable} summarizes our results. We observe that under both 
training conditions, our model achieves comparable performance. This indicates 
that our weakly supervised model can recover some of the information
that is missing while performing well at the activity categorization task.
In spite of using less
supervision at training time, our method outperforms state-of-the-art
methodologies that are trained with full spatial supervision.
% of the body parts
%and regions that are involved in each atomic actions.

\begin{table}[tb]
\footnotesize
\centering
\begin{tabular}{|l|c|}
\hline
\textbf{Algorithm} & \textbf{Accuracy}\\
\hline
Base model + GC, GEO desc. only, spatial supervision & 88.5\%\\
Base model + GC, with spatial supervision &  91.8\% \\
Our full model, no spatial supervision (latent $\vec{v}$) & 91.1\%\\
%Ours, GEO+TRAJ &  91.1\% \\
%Ours, GEO+TRAJ, NI  & 91.8\% \\
%Ours, GEO+TRAJ, NI, VL-KM-ST   & 91.1\% \\
\hline
%J. Luo et al. \cite{luo2013group} & \textbf{96.7\%} \\
%Y. Zhu et al. \cite{zhu2013fusing} & 94.3\% \\
Lillo \etal \cite{Lillo2014} (without GC) & 85.7\% \\
%BoW & 74.1\%    \\
%HMM & 78.9\%  \\
Cao et al. \cite{cao2015spatio} & 79.0\% \\
%H-BoW & 82.4\%   \\
%2-lev-HIER & 83.8\%  \\
\hline
\end{tabular}
\caption{
\footnotesize
Recognition accuracy in the Composable Activities
dataset.}
%\vspace{-1mm}
\label{tab:composable}
\end{table}
 
\subsection{Action Recognition in RGB Videos}
Our experiments so far have evaluated the performance of our model
in the task of human action recognition in RGBD videos.
In this experiment, we explore the use of our model in the problem of human
action recognition in RGB videos. For this purpose, we use the sub-JHMDB
dataset \cite{Jhuang2013}, which focuses on videos depicting 12 actions and
where most of the actor body is visible in the image frames.
In our validation, we use the 2D body pose configurations provided by the
authors and compare against previous methods that also use them. Given that 
this dataset only includes 2D image coordinates for each body joint, we obtain 
the geometric descriptor by adding a depth coordinate with a value $z = d$ to 
joints corresponding to wrist and knees, $z = -d$ to elbows, and $z = 0$ to other joints, 
so we can compute angles between segments, using $d = 30$ fixed with cross-validation. We summarize the results in Table 
\ref{tab:subjhmdb},
which shows that our method outperforms alternative state-of-the-art techniques.

%\paragraph{sub-JHMDB} In this dataset, we use the annotated joints provided to
%build our geometric descriptor. Only 15 joints per frame are annotated, and the
%coordinates of the joins are in 2D image coordinates.
%We first translate the
%15 joints into 20 joints, and also create a \emph{pseudo} 3D data by adding a
%$z=0$ coordinate to the joints, adding $d$ to the joints of wrists and knees,
%and subtracting $d$ for elbows, to create a 3D skeleton suitable to our model.
%AS RGB videos are available, we compute the TRAJ feature as in Composble
%Acivities Dataset (explain better).

%For a fair
%comparison, we compare our method with works that used the ground truth joints.
%We show in the results the mean accuracy over three splits, provided in the
%dataset.

\begin{table}[tb]
\footnotesize
\centering
\begin{tabular}{|l|c|}
\hline
\textbf{Algorithm} & \textbf{Accuracy}\\
\hline
Our model &  77.5\% \\
\hline
Huang et al. \cite{Jhuang2013} & 75.6\% \\
Ch\'eron et al. \cite{Cheron2015} & 72.5\%\\
\hline
\end{tabular}
\caption{\footnotesize
Recognition accuracy in the sub-JHMDB dataset.}
\label{tab:subjhmdb}
%\vspace{-1mm}
\end{table}

\subsection{Spatio-temporal Annotation of Atomic Actions}
In this experiment, we study the ability of our model to provide spatial and 
temporal annotations of relevant atomic actions. Table \ref{tab:annotation} 
summarizes our results. We report precision-recall rates
for the spatio-temporal annotations predicted by our model in the 
testing videos (first and second rows). Notice that this is a 
very challenging task. The testing videos do no provide any label, and 
the model needs to predict both, the temporal extent of each action and the 
body regions associated with the execution of each action. Although the 
difficulty of the task, our model shows satisfactory results being able to 
infer suitable spatio-temporal annotations.  

We also study the capability of the model to provide spatial and temporal 
annotations during training. In our first experiment, each video 
is provided
with the temporal extent of each action, so the model only needs to infer the 
spatial annotations (third row in Table \ref{tab:annotation}). In a 
second experiment, we do not provide any temporal or spatial annotation, 
but only the global action label of each video (fourth row in Table 
\ref{tab:annotation}). In both experiments, we observe that the model is 
still able to infer suitable spatio-temporal annotations.
%Note that this is due to the fact
%that we are not directly optimizing for atomic action annotation
%We also analyze the performance of the model in discovering the true atomic
%action annotations in the training set.
%our model.

\begin{table}[tb]
\footnotesize
\centering
\begin{tabular}{|l|c|c|c|}
\hline
\textbf{Videos} & \textbf{Annotation inferred} & \textbf{Precision} & \textbf{Recall}\\
\hline
Testing set & Spatio-temporal, no GC   & 0.59 & 0.77 \\
Testing set & Spatio-temporal   & 0.62 & 0.78 \\
\hline
Training set & Spatial only & 0.86 & 0.90\\
Training set & Spatio-temporal & 0.67 & 0.85 \\
\hline
\end{tabular}
\caption{
\footnotesize
Atomic action annotation performances in the Composable Activities
dataset. The results show that our model is able to recover spatio-temporal
annotations both at training and testing time.}
\label{tab:annotation}
%\vspace{-1mm}
\end{table}

\subsection{Effect of Model Components}
In this experiment,
we study the contribution of key components of the
proposed model. First, using the sub-JHMDB dataset, 
we measure the impact of three components of our model: garbage collector for 
motion poselets (GC), multimodal modeling of actionlets, and use of latent 
variables to infer spatial annotation about body regions (latent $\vec{v}$). Table 
\ref{tab:components} summarizes our experimental results. 
Table \ref{tab:components} shows that the full version
of our model achieves the best performance, with each of the components 
mentioned above contributing to the overall success of the method.

%In this dataset is clear tee benefits of all the components of the model: as we
%only have a single action label per video, we use the initialization of videos
%to get a better representation of the actions in the videos.
%As this dataset is
%from videos \emph{on the wild}, the camera view varies from video to video,
%making this dataset specially suitable to our algorithm of multiple classifiers
%per semantic action. Finally, the same as the rest of datasets, including the
%garbage collector math in the model allows to get a more discriminative model
%as it feeds the pose classifiers only with most informative poses.

\begin{table}[tb]
\footnotesize
\centering
\begin{tabular}{|l|c|}
\hline
\textbf{Algorithm} & \textbf{Accuracy}\\
\hline
Base model, GEO descriptor only & 66.9\%\\
Base Model & 70.6\%\\
Base Model + GC & 72.7\% \\
Base Model + Actionlets & 75.3\%\\
Our full model (Actionlets + GC + latent $\vec{v}$) &  77.5\% \\
\hline
\end{tabular}
\caption{
\footnotesize
Analysis of contribution to recognition performance from
each model component in the sub-JHMDB dataset.}
\label{tab:components}
%\vspace{-1mm}
\end{table}

Second, using the Composable Activities dataset, we also analyze the 
contribution of the proposed self-paced learning scheme for initializing and 
training our model. We summarize our results in
Table \ref{tab:initialization} by reporting action
recognition accuracy under different initialization schemes: i) Random: random 
initialization of latent variables $\vec{v}$, ii) Clustering: initialize 
$\vec{v}$ by first computing a BoW descriptor for the atomic action intervals 
and then perform $k$-means clustering, assigning the action intervals to the 
closer cluster center, and iii) Ours: initialize $\vec{v}$ using the proposed 
self-paced learning scheme. Our proposed initialization scheme helps the model to achieve its best
performance.

\begin{table}[tb]
\footnotesize
\centering
\begin{tabular}{|l|c|}
\hline
\textbf{Initialization Algorithm} & \textbf{Accuracy}\\
\hline
Random   & 46.3\% \\
Clustering   & 54.8\% \\
Ours   & 91.1\% \\
\hline
Ours, fully supervised & 91.8\%\\
\hline
\end{tabular}
\caption{
\footnotesize
Results in Composable Activities dataset, with latent $\vec{v}$ and different initializations. }
\label{tab:initialization}
%\vspace{-1mm}
\end{table}

\subsection{Qualitative Results}
\begin{figure}[tb]
\begin{center}
%\includegraphics[width=0.9\linewidth]{./Fig/poselets.pdf}
\scriptsize
 Motion poselet \#4 - most likely action: talking on cellphone\\
 \includegraphics[trim=0 0 0 0.35cm, clip, width=0.49\textwidth]{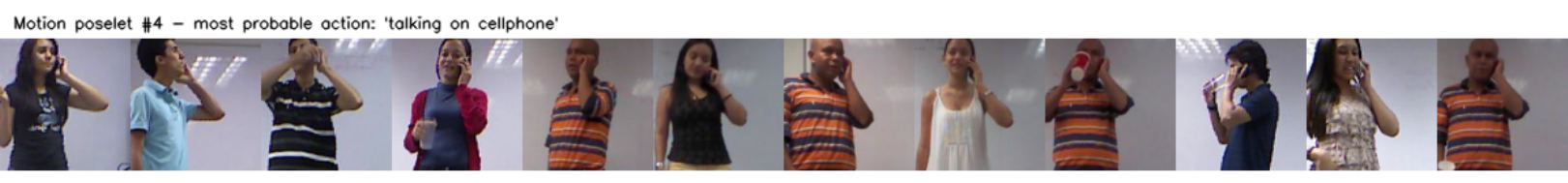}

 Motion poselet \#7 - most likely action: erasing on board\\
 \includegraphics[trim=0 0 0 0.35cm, clip, width=0.49\textwidth]{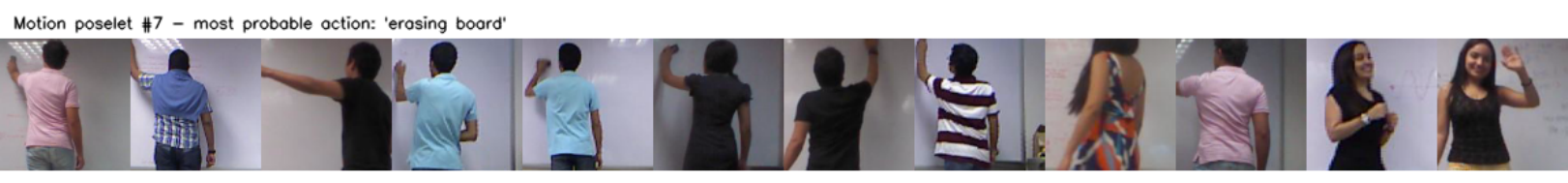}

 Motion poselet \#19 - most likely action: waving hand\\
 \includegraphics[trim=0 0 0 0.35cm, clip, width=0.49\textwidth]{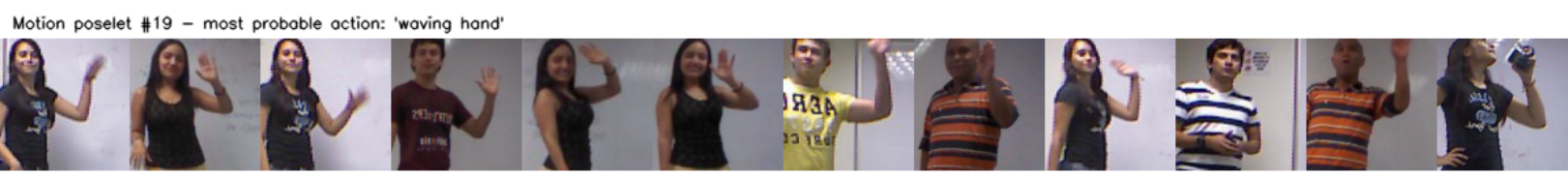}
\end{center}
%\vspace{-1mm}
\caption{
\footnotesize
Moving poselets learned from the Composable Activities
dataset.}
\label{fig:poselets_img}
\end{figure}

\begin{figure}[tb]
\begin{center}
\scriptsize
 Motion poselet \#16 - most likely action: tennis swing\\
 \includegraphics[trim=0 0 0cm 0cm, clip, width=0.49\textwidth]{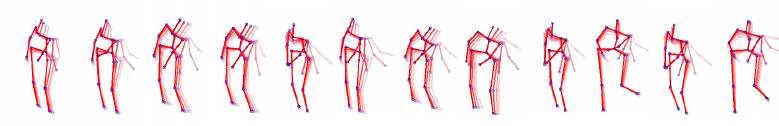}

 Motion poselet \#34 - most likely action: golf swing\\
 \includegraphics[trim=0 0 0cm 0cm,clip, width=0.49\textwidth]{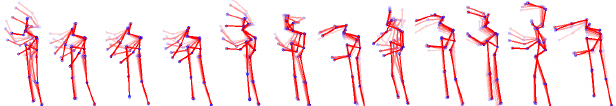}

 Motion poselet \#160 - most likely action: bend\\
 \includegraphics[trim=0 0 0cm 0cm, clip, width=0.49\textwidth]{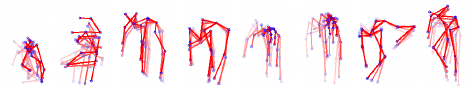}

\end{center}
%\vspace{-1mm}
\caption{
\footnotesize
Moving poselets learned from the MSR-Action3D
dataset.}
\label{fig:poselets_skel}
\end{figure}

Finally, we provide a qualitative analysis of
relevant properties of our model. Figure \ref{fig:poselets_img} 
shows examples of moving poselets learned in the Composable 
Activities dataset. We observe that each moving poselet captures 
a salient body configuration that helps to discriminate among atomic 
actions. To further illustrate this, Figure \ref{fig:poselets_img} 
indicates the most likely underlying atomic action for each moving poselet.
Figure \ref{fig:poselets_skel} presents a similar analysis for moving 
poselets learned in the MSR-Action3D dataset.

\begin{figure}[tb]
\begin{center}
\scriptsize
\includegraphics[]{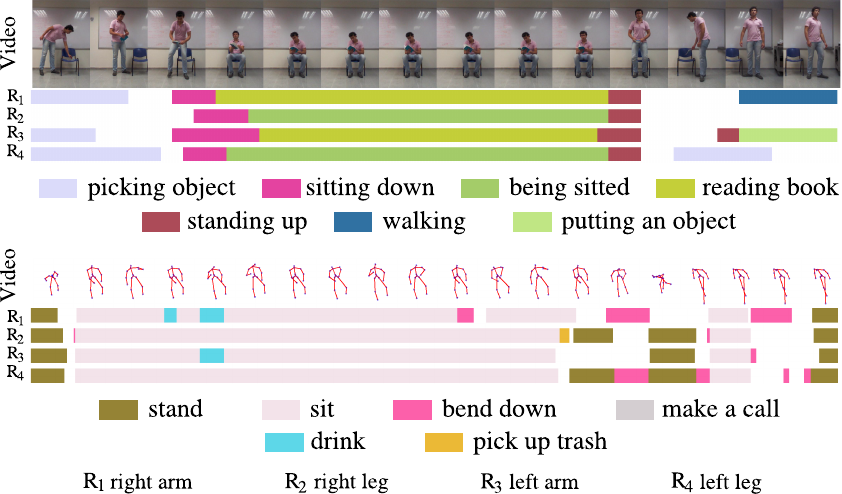}
\end{center}
%\vspace{-1mm}
\caption{
\footnotesize
Automatic spatio-temporal annotation of atomic actions. Our method
detects the temporal span and spatial body regions that are involved in
the performance of atomic actions in videos.}
\label{fig:actionlabels}
\end{figure}

We also visualize the action annotations produced by our model.
Figure \ref{fig:actionlabels} (top) shows the action labels associated
with each body part in a video from the Composable Activities dataset.
Figure \ref{fig:actionlabels} (bottom) illustrates per-body part action
annotations for a video in the Concurrent Actions dataset. These
examples illustrate the capabilities of our model to correctly
annotate the body parts that are involved in the execution of each action,
in spite of not having that information during training.

%We refer the reader to our supplementary material for additional technical
%details and example results.

\section{Conclusions and Future Work} \label{sec:conclusions} We present a hierarchical model for 
human action recognition using body joint locations. By using a semisupervised approach to jointly 
learn dictionaries of motions poselets and actionlets, the model demonstrates to be very flexible 
and informative, to handle visual variations and to provide spatio-temporal annotations of 
relevant atomic actions and active body part configurations. In particular, the model demonstrates 
to be competitive with respect to state-of-the -art approaches for complex action recognition, 
while also proving highly valuable additional information. As future work, the model can be 
extended to handle multiple actor situations, to use contextual information such as relevant 
objects, and to identify novel complex actions not present in the training set.

{\small
\textbf{Acknowledgements}
\label{sec:acknowledgements}
This work was partially funded by the FONDECYT grant 1151018, from CONICYT, Government of Chile;
and by the Stanford AI Lab-Toyota Center for Artificial Intelligence Research.
I.L. is supported by a PhD studentship from CONICYT.

}

{\small
\bibliographystyle{ieee}
\bibliography{AsaReferences_no_url,references,library_niebles_no_url,ILReferences_no_url,ivan}

\begin{thebibliography}{10}\itemsep=-1pt

\bibitem{Aggarwal2011}
J.~K. Aggarwal and M.~S. Ryoo.
\newblock {Human activity analysis}.
\newblock {\em ACM Computing Surveys}, 43(3):16:1--16:43, Apr. 2011.

\bibitem{Bourdev:EtAl:2010}
L.~Bourdev, S.~Maji, T.~Brox, and J.~Malik.
\newblock Detecting people using mutually consistent poselet activations.
\newblock In {\em ECCV}, pages 168--181, 2010.

\bibitem{Bourdev2009}
L.~Bourdev and J.~Malik.
\newblock Poselets: Body part detectors trained using {3D} human pose
  annotations.
\newblock In {\em {ICCV}}, pages 1365--1372, 2009.

\bibitem{cao2015spatio}
C.~Cao, Y.~Zhang, and H.~Lu.
\newblock Spatio-temporal triangular-chain crf for activity recognition.
\newblock In {\em Proceedings of the 23rd Annual ACM Conference on Multimedia
  Conference}, pages 1151--1154. ACM, 2015.

\bibitem{Cheron2015}
G.~Ch{\'{e}}ron, I.~Laptev, and C.~Schmid.
\newblock {P-CNN: Pose-based CNN Features for Action Recognition}.
\newblock {\em ICCV}, pages 3218--3226, 2015.

\bibitem{YongDu2015}
Y.~Du, W.~Wang, and L.~Wang.
\newblock {Hierarchical Recurrent Neural Network for Skeleton Based Action
  Recognition}.
\newblock In {\em CVPR}, pages 1110--1118, 2015.

\bibitem{Escorcia2012}
V.~Escorcia, M.~A. Davila, M.~Golparvar-Fard, and J.~C. Niebles.
\newblock Automated vision-based recognition of construction worker actions for
  building interior construction operations using {RGBD} cameras.
\newblock In {\em Construction Research Congress}, pages 879--888, 2012.

\bibitem{Feng2002}
X.~Feng and P.~Perona.
\newblock {Human action recognition by sequence of movelet codewords}.
\newblock In {\em 3DPVT}, volume~16, pages 717--721. IEEE, 2002.

\bibitem{Hu2015}
J.-F. Hu, W.-S. Zheng, J.~Lai, and J.~Zhang.
\newblock {Jointly learning heterogeneous features for RGB-D activity
  recognition}.
\newblock In {\em CVPR}, pages 5344--5352, 2015.

\bibitem{Hu2014}
N.~Hu, G.~Englebienne, Z.~Lou, and B.~Krose.
\newblock {Learning latent structure for activity recognition}.
\newblock In {\em ICRA}, 2014.

\bibitem{Jhuang2013}
H.~Jhuang, J.~Gall, S.~Zuffi, C.~Schmid, and M.~J. Black.
\newblock {Towards understanding action recognition}.
\newblock In {\em ICCV}, pages 3192--3199, 2013.

\bibitem{Joachims2009}
T.~Joachims, T.~Finley, and C.~Yu.
\newblock Cutting-plane training of structural {SVM}s.
\newblock {\em Machine Learning}, 77(1):27--59, 2009.

\bibitem{Johansson:1973}
G.~Johansson.
\newblock Visual perception of biological motion and a model for its analysis.
\newblock {\em Perception \& Psychophysics}, 14(2):201--211, 1973.

\bibitem{Kong2015}
Y.~Kong and Y.~Fu.
\newblock {Bilinear heterogeneous information machine for RGB-D action
  recognition}.
\newblock In {\em CVPR}, pages 1054--1062, 2015.

\bibitem{Kumar:EtAl:2010}
M.~P. Kumar, B.~Packer, and D.~Koller.
\newblock Self-paced learning for latent variable models.
\newblock In {\em NIPS}, pages 1189--1197, 2010.

\bibitem{Lan2015}
T.~Lan, Y.~Zhu, A.~R. Zamir, and S.~Savarese.
\newblock Action recognition by hierarchical mid-level action elements.
\newblock In {\em ICCV}, pages 4552--4560, 2015.

\bibitem{WanLi2010}
W.~Li, Z.~Zhang, and Z.~Liu.
\newblock Action recognition based on a bag of {3D} points.
\newblock In {\em {CVPR}}, pages 9--14, 2010.

\bibitem{Lillo2014}
I.~Lillo, A.~Soto, and J.~C. Niebles.
\newblock Discriminative hierarchical modeling of spatio-temporally composable
  human activities.
\newblock In {\em {CVPR}}, pages 812--819, 2014.

\bibitem{maji2011action}
S.~Maji, L.~Bourdev, and J.~Malik.
\newblock Action recognition from a distributed representation of pose and
  appearance.
\newblock In {\em CVPR}, pages 3177--3184, 2011.

\bibitem{Nie2015}
B.~X. Nie, C.~Xiong, and S.-c. Zhu.
\newblock {Joint action recognition and pose estimation from video}.
\newblock In {\em CVPR}, pages 1293--1301, 2015.

\bibitem{Poppe2010}
R.~Poppe.
\newblock {A survey on vision-based human action recognition}.
\newblock {\em Image and Vision Computing}, 28(6):976--990, 2010.

\bibitem{Ramanan2003}
D.~Ramanan and D.~A. Forsyth.
\newblock {Automatic annotation of everyday movements}.
\newblock In {\em {NIPS}}, 2003.

\bibitem{Raptis2012}
M.~Raptis, I.~Kokkinos, and S.~Soatto.
\newblock {Discovering discriminative action parts from mid-level video
  representations}.
\newblock {\em CVPR}, pages 1242--1249, 2012.

\bibitem{Rohrbach2012}
M.~Rohrbach, S.~Amin, M.~Andriluka, and B.~Schiele.
\newblock {A database for fine grained activity detection of cooking
  activities}.
\newblock In {\em CVPR}, pages 1194--1201, 2012.

\bibitem{Shotton:EtAl:11}
J.~Shotton, A.~Fitzgibbon, M.~Cook, T.~Sharp, M.~Finocchio, R.~Moore,
  A.~Kipman, and A.~Blake.
\newblock Real-time human pose recognition in parts from a single depth image.
\newblock In {\em Communications of the {ACM}}, pages 116--124, 2011.

\bibitem{Tao2015}
L.~Tao and R.~Vidal.
\newblock {Moving Poselets : A Discriminative and Interpretable Skeletal Motion
  Representation for Action Recognition}.
\newblock In {\em {IEEE} International Conference on Computer Vision
  Workshops}, pages 61--69, 2015.

\bibitem{Thurau2008}
C.~Thurau and V.~Hlavac.
\newblock {Pose primitive based human action recognition in videos or still
  images}.
\newblock In {\em {CVPR}}, pages 1--8, 2008.

\bibitem{Vemulapalli2014}
R.~Vemulapalli, F.~Arrate, and R.~Chellappa.
\newblock {Human Action Recognition by Representing 3D Skeletons as Points in a
  Lie Group}.
\newblock In {\em CVPR}, pages 588--595, 2014.

\bibitem{vishwakarma2013survey}
S.~Vishwakarma and A.~Agrawal.
\newblock A survey on activity recognition and behavior understanding in video
  surveillance.
\newblock {\em The Visual Computer}, 29(10):983--1009, 2013.

\bibitem{Wang2013}
C.~Wang, Y.~Wang, and A.~L. Yuille.
\newblock {An approach to pose-based action recognition}.
\newblock In {\em {CVPR}}, pages 915--922, 2013.

\bibitem{WangCVPR2011}
H.~Wang, A.~Klaser, C.~Schmid, and C.-L. Liu.
\newblock Action recognition by dense trajectories.
\newblock In {\em {CVPR}}, pages 3169--3176, 2011.

\bibitem{Wang2012}
J.~Wang, Z.~Liu, Y.~Wu, and J.~Yuan.
\newblock Mining actionlet ensemble for action recognition with depth cameras.
\newblock In {\em {CVPR}}, pages 1290--1297, 2012.

\bibitem{Wang2014}
L.~Wang, Y.~Qiao, and X.~Tang.
\newblock {Video Action Detection with Relational Dynamic-Poselets}.
\newblock In {\em ECCV}, pages 565--580, 2014.

\bibitem{Wang2008}
Y.~Wang and G.~Mori.
\newblock {Learning a discriminative hidden part model for human action
  recognition}.
\newblock In {\em {NIPS}}, pages 1721--1728, 2008.

\bibitem{Wei2013}
P.~Wei, N.~Zheng, Y.~Zhao, and S.-C. Zhu.
\newblock Concurrent action detection with structural prediction.
\newblock In {\em {ICCV}}, pages 3136--3143, 2013.

\bibitem{weinland2011survey}
D.~Weinland, R.~Ronfard, and E.~Boyer.
\newblock A survey of vision-based methods for action representation,
  segmentation and recognition.
\newblock {\em Journal of Computer Vision and Image Understanding},
  115(2):224--241, 2011.

\bibitem{Yao2010}
B.~Yao and L.~Fei-Fei.
\newblock {Modeling Mutual Context of Object and Human Pose in Human-Object
  Interaction Activities}.
\newblock In {\em {CVPR}}, pages 17--24. IEEE, 2010.

\bibitem{Yeung2015}
S.~Yeung, O.~Russakovsky, N.~Jin, M.~Andriluka, G.~Mori, and L.~Fei-Fei.
\newblock {Every Moment Counts: Dense Detailed Labeling of Actions in Complex
  Videos}.
\newblock {\em arXiv:1507.05738}, 2015.

\bibitem{Yu:Joachims:2010}
C.~Yu and T.~Joachims.
\newblock Learning structural svms with latent variables.
\newblock In {\em ICML}, pages 1169--1176, 2009.

\bibitem{Yuille:Rangarajan:03}
A.~Yuille and A.~Rangarajan.
\newblock The concave-convex procedure.
\newblock {\em Neural Computation}, 15(4):915--936, 2003.

\bibitem{Zanfir2013}
M.~Zanfir, M.~Leordeanu, and C.~Sminchisescu.
\newblock {The Moving Pose: An Efficient 3D Kinematics Descriptor for
  Low-Latency Action Recognition and Detection}.
\newblock In {\em ICCV}, pages 2752--2759, 2013.

\end{thebibliography}
}

\end{document}